\documentclass[11pt]{article}

\usepackage[preprint]{acl}

\usepackage{times}
\usepackage{latexsym}

\usepackage[T1]{fontenc}

\usepackage[utf8]{inputenc}

\usepackage{microtype}

\usepackage{inconsolata}

\usepackage{graphicx}

\usepackage{url}            
\usepackage{booktabs}       
\usepackage{amsfonts}       
\usepackage{nicefrac}       
\usepackage{microtype}      
\usepackage{xcolor}         
\usepackage{graphicx}
\usepackage{subcaption}
\usepackage{wrapfig}
\usepackage{lipsum}
\usepackage{amsmath}
\usepackage{multirow}
\usepackage{xspace}
\usepackage{algorithm}
\usepackage{algpseudocode}
\usepackage{tikz}
\newcommand*\circled[1]{\tikz[baseline=(char.base)]{
    \node[shape=circle,draw,inner sep=0.02pt] (char) {#1};}}

\newcommand{\inserthyperparams}{
	\begin{table}[h]
		\centering
		\scriptsize
		\begin{tabular}{lcccc}
        \toprule
        \textbf{Model} & \textbf{Encoder} & \textbf{LM Module} & \textbf{Decoder NB} & \textbf{Decoder MB} \\
         &  & \textbf{Module} & \textbf{NT} & \textbf{MB} \\
		\midrule
		LlamaByte & 22 & 0 & 0 & 0 \\
		SpaceByte & 2 & 16 & 4 & 0 \\
		FxT & 2 & 16 & 4 & 0 \\
		MLP-MBP & 2 & 16 & 2 & 3 \\
		\methodname-MBP & 2 & 16 & 2 & 2 \\
		\bottomrule
		\end{tabular}
		\caption{Architectural configurations used across model variants. Unless otherwise stated, all models use the same hidden dimension, training hyperparameters, and optimizer settings. For MLP-MBP trained with compression rate $5$, the MB decoder uses $5$ layers instead of $3$.}
		\label{tab:hyperparams}
	\end{table}
}

\newcommand{\insertacceptancerate}{
    \begin{table}[h]
        \centering
        \resizebox{\columnwidth}{!}{
        \begin{tabular}{lccccc}
        \toprule
        \textbf{Model} & \textbf{IfEval} & \textbf{CoQA} & \textbf{DailySum} & \textbf{es-en} & \textbf{fr-en} \\
        \midrule
        MLP-MBP & 61.91 & 63.45 & 58.08 & 62.67 & 65.01 \\
        \methodname-MBP & 45.95 & 46.62 & 48.89 & 51.68 & 49.74 \\
        \bottomrule
        \end{tabular}
        }
        \caption{Byte acceptance rates of MLP and \methodname multi-byte prediction heads across five downstream tasks. Acceptance rate measures the fraction of predicted bytes accepted. While MLP achieves higher raw acceptance, indicating that the model is highly confident about its prediction, it performs worse than \methodname-MBP on the Pareto front.}
        \label{tab:acceptance_rate}
        \vspace{-2em}
    \end{table}
}
\newcommand{\insertnablationthroughput}{
    \begin{table}[h]
        \centering
        \small
        \begin{tabular}{lccc}
        \toprule
        $n$ & \textbf{DailySum} & \textbf{es-en} & \textbf{fr-en} \\
        \midrule
        3 & 210.42 & 285.75 & 278.21 \\
        4 & 205.42 & 283.47 & 278.13 \\
        5 & 194.50 & 254.57 & 263.99 \\
        6 & 181.24 & 234.55 & 252.31 \\
        7 & 164.18 & 221.09 & 230.99 \\
        \bottomrule
        \end{tabular}
        \caption{Throughput (bytes/sec) of \methodname-MBP across speculative-candidate counts $n \in \{3, 4, 5, 6, 7\}$ at acceptance threshold $\tau=0.75$. Throughput decreases monotonically with $n$ on all three tasks, as larger speculative windows incur more rejected candidates per step.}
        \label{tab:n_ablation_throughput}
    \end{table}
}

\newcommand{\insertverifyallacceptancerate}{
    \begin{table}[h]
        \centering
        \small
        \begin{tabular}{lccc}
        \toprule
        \textbf{Model} & \textbf{DailySum} & \textbf{es-en} & \textbf{fr-en} \\
        \midrule
        FxT              & 100.00 & 100.00 & 100.00 \\
        LCA NBP          & 100.00 & 100.00 & 100.00 \\
        LCA Self-verify     & 46.36	& 45.85	 & 45.85 \\
        LCA FxT-verify   & 47.97 & 50.77 & 52.04 \\
        \bottomrule
        \end{tabular}
        \caption{Token acceptance rates (\%) across three tasks under self and external (FxT) verification using $n=7$ candidate bytes.}
        \label{tab:verify_all_acceptance_rate}
    \end{table}
}

\newcommand{\insertspecdecodeperf}{
    \begin{table}[h]
        \centering
        \resizebox{\columnwidth}{!}{%
        \begin{tabular}{lcccccc}
        \toprule
        \textbf{Model} & \textbf{IfEval} & \textbf{CoQA} & \textbf{DailySum} & \textbf{es-en} & \textbf{fr-en} & \textbf{Avg} \\
        \midrule
        \multicolumn{7}{l}{\textit{Performance}} \\
        \midrule
        MLP-MBP  & 46.54 & 20.85 & 32.73 & 0.8178 & 0.8064 & -- \\
        LCA-MBP  & 44.97 & 22.11 & 38.09 & 0.8147 & 0.7977 & -- \\
        \midrule
        \multicolumn{7}{l}{\textit{Acceptance rate (\%)}} \\
        \midrule
        MLP-MBP  & 71.80 & 90.15 & 90.79 & 91.82 & 91.82 & 87.28 \\
        LCA-MBP  & 51.17 & 73.93 & 81.22 & 79.85 & 79.85 & 73.20 \\
        \bottomrule
        \end{tabular}%
        }
        \caption{Performance and acceptance rate under speculative decoding acceptance. Performance is constant across the number of speculative candidates $n$; acceptance rates are reported at $n=3$.}
        \label{tab:spec_decode_performance}
    \end{table}
}

\newcommand{\methodname}{\textsc{LCA}\xspace}

\newcommand{\Sref}[1]{\S\ref{#1}}

\title{Dynamic Multi-Byte Prediction With Hierarchical Language Models}

\author{Abraham Toluwase Owodunni\textsuperscript{1} \ \ Chibuzor Okocha\textsuperscript{2} \ \ Christan Grant\textsuperscript{2} \ \ Tomasz Limisiewicz\textsuperscript{3} \\ \textbf{Sachin Kumar}\textsuperscript{1}\\
\textsuperscript{1}The Ohio State University \ \ \textsuperscript{2}University of Florida \ \
\textsuperscript{3} University of Washington \\
\href{mailto:owodunni.1@osu.edu}{\texttt{owodunni.1@osu.edu}}
}

\begin{document}
\maketitle
\begin{abstract}

Byte-level hierarchical language models (LMs) have recently emerged as a robust alternative to their popular counterparts that use subword tokenization. However, generating one byte at a time remains a bottleneck for inference speed. To address this, we introduce multi-byte prediction (MBP) for hierarchical models, which generates multiple bytes in parallel, speeding up inference with minimal performance impact and no additional parameters.
MBP builds on the popular multi-token prediction (MTP) paradigm with two crucial innovations. First, we introduce a variable-length prediction window that aligns with the latent tokens, or segments, of a hierarchical LM. Second, we implement a novel attention-masking scheme that enables parallel byte prediction without violating causality. 
We show that multi-byte prediction strikes a Pareto-optimal trade-off across multiple generative tasks, instruction following, question answering, summarization, and machine translation, achieving the best trade-off between performance and inference throughput.  Code  for our experiments will be released at \href{https://github.com/skai-research/lca-multibyte}{github.com/skai-research/lca-multibyte}.

\end{abstract}

\section{Introduction}

Subword tokenization has been a central design choice in modern language models, enabling efficient text compression and open-vocabulary generation \cite{sennrich-etal-2016-neural, kudo2018subword}. However, subword tokenizers, due to their predetermined fixed vocabulary, may overfragment rare words, encode language- and script-specific biases, and generalize poorly across domains, non-English settings, and evolving scenarios \cite{mielke2021between, menschikov2025beyond}. Byte-level language modeling \citep{xue-etal-2022-byt5, wang2024mambabyte,yu2023megabyte} offers an appealing alternative: by operating directly on raw bytes, models can eliminate tokenizer-specific assumptions and represent any text sequence. 
\begin{figure}[t]
    \centering
    \includegraphics[width=\linewidth]{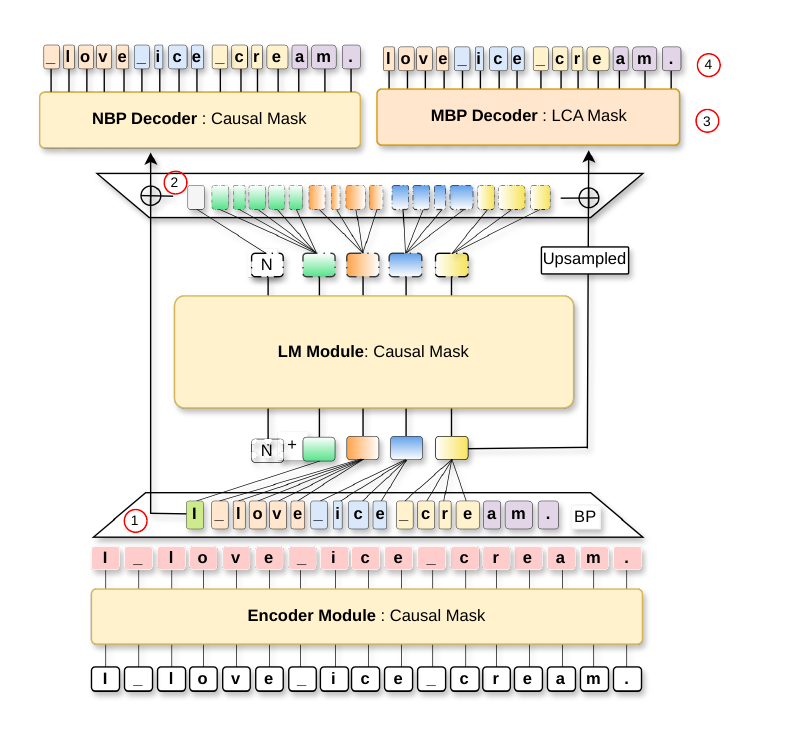}
    \vspace{-0.8cm}
    \caption{\textbf{Overview of the MBP architecture.} An input byte sequence is first processed by the model, and the MBP head uses the \methodname mask (\autoref{fig:gca_mask}) to predict multiple future bytes in parallel from a single head. Colored bands denote learned segment boundaries. See \autoref{apx:archi_and_hyperparams} for a detailed description of several parts in this figure.}
    \label{fig:mbp_lca_architecture}
    \vspace{-0.5cm}
\end{figure}

Despite these advantages, byte-level modeling remains computationally expensive \cite{tay2022charformer,yu2023megabyte}. Byte sequences are substantially longer than subword sequences, which increases the cost of Transformer attention and makes autoregressive decoding slower \cite{xue-etal-2022-byt5, wang2024mambabyte, slagle2024spacebyte}. 
Recent hierarchical byte-level models attempt to address this issue by compressing byte streams into shorter latent sequences \cite{slagle2024spacebyte, pagnoni2025byte, hwang2025dynamic, owodunni2025flexitokens} by incorporating heuristic or learned tokenization mechanisms that reduce the effective sequence length before applying deeper Transformer computation. These approaches show that byte-level models benefit from operating over higher-level units rather than individual bytes alone. However, most hierarchical designs use latent tokenization primarily for representation efficiency: the model compresses the input, processes the shorter sequence, and then decodes back to bytes \cite{hwang2025dynamic, neitemeier2025hierarchical}. Inference still happens one byte at a time.


In parallel, multi-token prediction (MTP) has emerged as a promising approach for accelerating language model decoding. Early work introduced the idea of predicting multiple future tokens in a single forward pass to reduce decoding latency \citep{stern2018blockwise, qi-etal-2020-prophetnet}, and more recent methods have shown that this objective also improves sample efficiency at scale \citep{gloeckle2024better}. Existing MTP approaches typically rely on multiple prediction heads attached to the model trunk \citep{cai2024medusa, ankner2024hydra} or on smaller draft models that propose candidates for the main model to verify \citep{leviathan2023fast, chen2023accelerating, li2024eagle}. While these designs can improve inference speed, they have two important limitations. First, predicting more future tokens typically requires additional heads or auxiliary parameters for every token, increasing memory cost. Second, fixed-offset prediction treats the next $n$ tokens as a static horizon, ignoring the fact that language has variable local structure: some regions are easily predictable as coherent chunks, while others require finer-grained sequential modeling.

This paper proposes Latent Causal Attention (\methodname), a dynamic multi-\textit{byte} prediction method for hierarchical language models. Our key idea involves the use of learned byte segments not only for compression, but also as the unit of multi-byte generation. Instead of assigning one prediction head to each future byte, \methodname uses a single multi-byte decoder equipped with a boundary-aware causal mask that allows each byte in a segment to attend to previous segments while preventing dependence on other bytes within the same segment. This preserves global autoregressive causality while enabling all bytes in a predicted segment to be generated in parallel. This design replaces the fixed-offset prediction of prior MTP methods with variable-length, segment-aligned generation. \methodname removes the need for additional prediction heads as the number of generated tokens increases and provides a natural extension of hierarchical byte-level modeling to inference acceleration.
We describe our methodology in detail in Section \ref{sec:method}. We then present our findings and extensive analyses in Sections \ref{sec:results} and \ref{sec:analysis}, respectively.

\section{Background}
\label{sec:background}
\subsection{Multi-token Prediction} 

Multi-token prediction (MTP) has emerged as a way to improve both sample efficiency and inference speed of autoregressive language models by predicting multiple future subword tokens in parallel, as opposed to one token at a time \cite{gloeckle2024better,cai2024medusa}.  Given a sequence $x_1, \dots, x_T$, MTP is trained by minimizing 
$\mathcal{L}_{\text{MTP}} = -\sum_{t} \log P_{\theta}(x_{t+1:t+n} \mid x_{1:t})$ 
where $\theta$ parameterizes the LM and
$n$ is the number of future tokens to be predicted.
To make this objective tractable, $n$ dedicated prediction heads (typically multiple MLP layers or independent transformer blocks followed by a shared unembedding matrix) are used to predict each of the $n$ future tokens independently and in parallel. Each head operates on a shared hidden representation $z_t$ encoding $x_{1:t}$. Under this conditional independence assumption, the joint probability factorizes and loss becomes:
\begin{align}
\mathcal{L}_{\text{MTP}} = -\sum_{t} \sum_{i=1}^{n} \log P_{\theta}(x_{t+i} \mid z_{t})
\label{eq:mtp_factorized}
\end{align}

Since different parts of a sequence exhibit varying levels of surprisal, always predicting a fixed $n$ tokens per forward pass makes this approach rigid and unable to adapt to the local difficulty of the text. Moreover, this design also introduces a parameter bottleneck: each additional future token requires its own prediction head, linearly increasing the prediction head's parameter count with $n$. 

\subsection{Hierarchical Language Models}

Hierarchical byte-level language models have recently emerged as a viable alternative to subword-based language models. In this work, we build on the architectural design of FlexiTokens \citep{owodunni2025flexitokens}, introducing multiple modifications to support dynamic multi-byte prediction. Our hierarchical model comprises four components (\autoref{fig:mbp_lca_architecture}): an encoder module, a language modeling (LM) module, a boundary predictor, and a decoder module. We note that our proposed method is generalizable to other hierarchical LM \citep{pagnoni2025byte,hwang2025dynamic} architectures since they follow a similar design.

\paragraph{Encoder Module.}
This module maps an input byte sequence $x_1, \ldots, x_T$ to hidden states $h_e \in \mathbb{R}^{T \times d}$ via transformer layers with a causal attention mask. $h_e$ serves both as a rich contextual representation and as a residual connection for downstream modules. We then feed $h_e$ into a boundary predictor $\mathcal{B}$, which produces boundary probabilities $\hat{B} \in [0, 1]^T$, indicating positions where a segment should end. A discrete label is sampled from these probabilities using Gumbel Sigmoid trick to maintain differentiability \citep{nawrot-etal-2023-efficient}. At positions where the label is 1, $h_e$ is downsampled via mean pooling with all preceding positions where the label is 0 to form a shortened hidden representation $h_{\downarrow}$ or \textit{latent tokens}, with sequence length strictly less than that of $h_e$.
The boundary predictor in this module is jointly optimized during training; we use a boundary predictor loss $\mathcal{L}_{\mathrm{BP}}$ that controls the compression of the input sequence. Following \citet{owodunni2025flexitokens}, we define:
\begin{align*}
    \mathcal{L}_{\mathrm{BP}} = \max\!\left(\frac{k}{T} - \alpha,\ 0\right) + \max\!\left(\beta - \frac{k}{T},\ 0\right)
\end{align*}
where $k$ is the number of boundaries predicted as 1.
Hyperparameters $\alpha$ and $\beta$ define upper and lower bounds on $k/T$, respectively, such that the compression rate is considered optimal when $\beta \leq \frac{k}{T} \leq \alpha$.

\paragraph{Language Modeling Module.}
A learned \texttt{[BOS]} token embedding is appended to $h_{\downarrow}$ before being passed through the LM module's transformer stack to output hidden vectors $h'_{\downarrow}$. The \texttt{[BOS]} token shifts the shortened sequence forward by one, ensuring that the representation at each latent token position conditions only on preceding tokens, thereby preserving causality. 
$h'_{\downarrow}$ is subsequently upsampled via duplication to match the original byte sequence length and passed to the decoder. 

\paragraph{Decoder Module.}
This module produces byte-level predictions. We add a residual connection $r_1 = h_e$ from the encoder to the upsampled output of the LM module. The result is passed through a small stack of transformer layers, followed by an unembedding layer forming the primary next-byte prediction head, optimized with cross-entropy loss. 
  \begin{figure}[t]
    \centering
    \includegraphics[width=\linewidth]{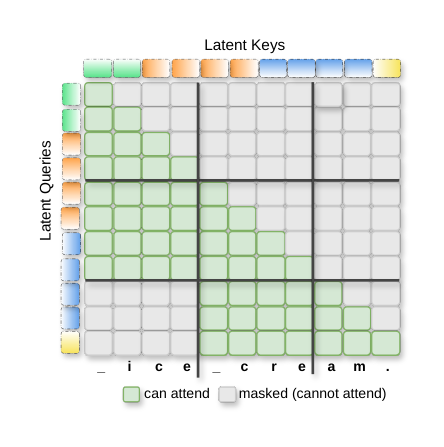}
    \caption{\textbf{The Latent Causal Attention (\methodname) mask.} Each row corresponds to a query byte and each column to a key byte. A query at position $i$ in segment $s_i$ may attend to itself, to earlier positions within its own segment, and to all bytes in the immediately preceding segment $s_{i-1}$. Attention to later positions in the same segment and to segments further in the past is masked out. Colored bands denote segment boundaries and match those in \autoref{fig:mbp_lca_architecture}. This mask preserves autoregressive causality at the segment level while enabling all bytes in a group to be predicted in parallel from a single decoder head.}
    \label{fig:gca_mask}
\end{figure}

\begin{algorithm*}[t]
\caption{Multi-byte Prediction with \methodname}
\label{alg:inference}
\begin{algorithmic}[1]
\Require Prompt $x_{1:t}$, boundary predictor $\mathcal{B}$, threshold $\tau$, \#MBP candidates $n$
\While{not done}
    \State $h_e \gets \mathrm{EncoderModule}(x_{1:t})$; cache residual $r_1 \gets h_e$
    \State $b \gets \mathcal{B}(h_e)$ \Comment{Predict boundary}
    \If{$b = 1$ \textbf{or} no cached $h_{\downarrow}$}
        \State $h_{\downarrow} \gets \mathrm{Downsample}(h_e)$; cache residual $r_2 \gets h_{\downarrow}$
        \State $h'_{\downarrow} \gets \mathrm{LMModule}(h_{\downarrow})$
    \EndIf
    \State $h_1 \gets \mathrm{NTLayers}(\mathrm{Upsample}(h'_{\downarrow}) + r_1)$
    \State $h_2 \gets \mathrm{Upsample}(h'_{\downarrow} + r_2)$ \Comment{Residual added \emph{before} upsample}
    \State $h_2 \gets \big[\,h_2 \,;\; \underbrace{h_2[-1],\,\ldots,\,h_2[-1]}_{n-2\text{ copies}}\,\big]$ \Comment{Duplicate last token $n{-}2$ times}
    \State $s \gets$ start index of \emph{previous} segment boundary
    \State $h_2 \gets \mathrm{MBPLayers}\!\big(h_2[s:],\; \mathrm{segment\;mask}\big)$ \Comment{Only prev.\ + current segment}
    \State $t_1 \sim \mathrm{Head}(h_1[-1])$ \Comment{NT: always accepted}
    \State $t_{2:n+1} \sim \mathrm{Head}(h_2[-(n-1):])$ \Comment{MBP: $n$ speculative}
    \State Accept $t_{2:k}$ while $P(t_k) \geq \tau$; discard rest
    \State Append accepted bytes to $x_{1:t}$
\EndWhile
\end{algorithmic}
\end{algorithm*}

\section{Multi-Byte Prediction with Latent Causal Attention}
\label{sec:method}
\label{sec:lca}
Here we describe our modifications to the hierarchical LM architecture to enable multi-byte prediction. 
We decouple the decoder module into only two heads: a standard next-byte prediction head (same as before) and a single multi-byte prediction head (MBP; \autoref{fig:mbp_lca_architecture} \circled{3}). The MBP head is trained to predict all bytes in a latent token in parallel, except the first which is predicted by the next-byte head. Since the number of bytes in each latent token is variable, a fixed number of independent prediction heads, typically used in the MTP literature, is impractical. 
Instead, the MBP head consists of transformer layers.

The input to the MBP head is the duplicated upsampled output of the LM module. We add a residual connection $r_2 = h_\downarrow$ from the downsampled representation to each of these vectors. 

Since the LM module shifts the latent tokens by one position, this input only carries information about the bytes from the previous latent token.  
For each transformer layer in the MBP head, 
we construct an attention mask (\autoref{fig:gca_mask}) such that each position may attend to itself, to earlier positions within the latent token, and to all bytes in the immediately preceding latent token. 
Intra-token attention does not violate causality at inference time, as the input to each position within the same token are duplicates plus residuals of the previous latent token.
We call this mask \textit{Latent Causal Attention} (\methodname) and our overall method as LCA-MBP.
The output of the transformer layers is multiplied by an unembedding matrix to predict byte logits over which the cross-entropy loss is computed to predict target bytes shifted by two positions   (\autoref{fig:mbp_lca_architecture} \circled{4}). Both decoder heads share a single unembedding matrix. 

The overall training objective combines the next-byte prediction loss with the multi-byte prediction loss:

\begin{multline}
\mathcal{L}_{\mathrm{LM}} = -\lambda_0 \sum_{t} \log P_{\theta}(x_{t+1} \mid x_{1:t}) \\
- \lambda_1 \sum_{t} \log P_{\theta}(x_{t+2} \mid x_{i:t})
\label{eq:gca_loss}
\end{multline}

where $i$ is the starting index of the $x_t$'s immediately previous latent token. $\lambda_0$ and $\lambda_1$ are hyperparameters. Unlike standard MTP, which requires separate prediction heads, \methodname{} achieves this independence through the latent attention mask and byte segments pooling, while using a single set of shared transformer layers. 
The final loss we train this model with is $\mathcal{L}_\mathrm{LM} + \lambda_2 \mathcal{L}_\mathrm{BP}$, where $\lambda_2$ is another hyperparameter.




\subsection{Inference Algorithm}
We present our inference methodology in Algorithm \autoref{alg:inference}. At inference time, the prompt $x_{1:t}$ is first passed through the encoder module to produce a hidden representation $h_e$, from which we store a residual connection $r_1 = h_e$. The boundary predictor is then applied to $h_e$ to identify boundaries. 

To generate the first token, if the boundary predicted at position $x_t$ is 1, or no cached LM module state exists, we downsample $h_e$ to obtain $h_{\downarrow}$ (Line~5), store a residual $r_2 = h_{\downarrow}$, and pass $h_{\downarrow}$ through the LM module to produce $h'_{\downarrow}$ (Line~6). Otherwise, the cached $h'_{\downarrow}$ from the previous step is reused, avoiding a full LM forward pass.

To predict the next byte, we add the residual $r_1$ to $h'_{\downarrow}$, and process the result through the next-byte head (Line~8). To predict the next $n$ bytes with the MBP head, we add the residual $r_2$ to $h'_{\downarrow}$ before upsampling to obtain $h_2$. We then slice $h_2$ to retain only the bytes belonging to the current and previous latent tokens, and pass this window through the MBP transformer layers with the \methodname mask (Lines~9--12).

\paragraph{Byte Acceptance.}
The output of the next-byte head is always accepted (Line~14). For the MBP head, a predicted token $\hat{x}_{t+i}$ is accepted only if the model's confidence exceeds a predefined threshold: 
\begin{align*}
    \hat{x}_{t+i} \text{ is accepted} \iff P_\theta(\hat{x}_{t+i} \mid \cdot) \geq \tau 
\end{align*}
Tokens are accepted from left to right; the first token falling below $\tau$ and all subsequent candidates are discarded. This acceptance strategy is applied uniformly across all MBP methods evaluated in this work, following \citet{kirchenbauer2026multi}.

\section{Experimental Setup}

\subsection{Baselines}
We compare \methodname against five baselines spanning flat and hierarchical byte-level architectures.

\begin{itemize}
    \item \textbf{LlamaByte} \citep{grattafiori2024llama}: a vanilla byte-level model implemented with the Llama 3 architecture.

    \item \textbf{SpaceByte} \cite{slagle2024spacebyte}: a hierarchical byte-level model that uses an external fixed heuristic for segmentation during training. An external fixed heuristic was also used in BLT \citep{pagnoni2025byte}.

    \item \textbf{FlexiTokens (FxT)} \citep{owodunni2025flexitokens, hwang2025dynamic}: a hierarchical model with a learned boundary predictor that dynamically segments byte sequences into variable-length segments. This boundary predictor is jointly optimized with the language modeling objective during training.

    \item \textbf{MLP-MBP} \citep{gloeckle2024better}: a multi-byte prediction baseline built on the FxT architecture, which employs $n$ independent MLP heads to predict $n$ future tokens in parallel, one per head, following the setup of \citet{gloeckle2024better} and Medusa \citep{cai2024medusa}.

    \item \textbf{Efficient FlexiTokens (Eff-FxT)}: a variant of FxT that uses a more efficient inference algorithm. Specifically, the LM module is only invoked at predicted boundary positions; at all other positions, the cached LM module output is reused. This corresponds exactly to Lines 4--8 in Algorithm \autoref{alg:inference}. We note that HNet \citep{hwang2025dynamic} uses a similar generation strategy in their work as well.
\end{itemize}

To ensure a fair comparison, all models are parameter-matched by scaling each baseline to the same total parameter count, which consequently increases the floating-point operations (FLOPs) of flat models such as LlamaByte and SpaceByte. See \autoref{tab:hyperparams} in \autoref{apx:archi_and_hyperparams} for the number of layers used across all methods.

\begin{figure*}[t]
    \centering
    \includegraphics[width=1\linewidth]{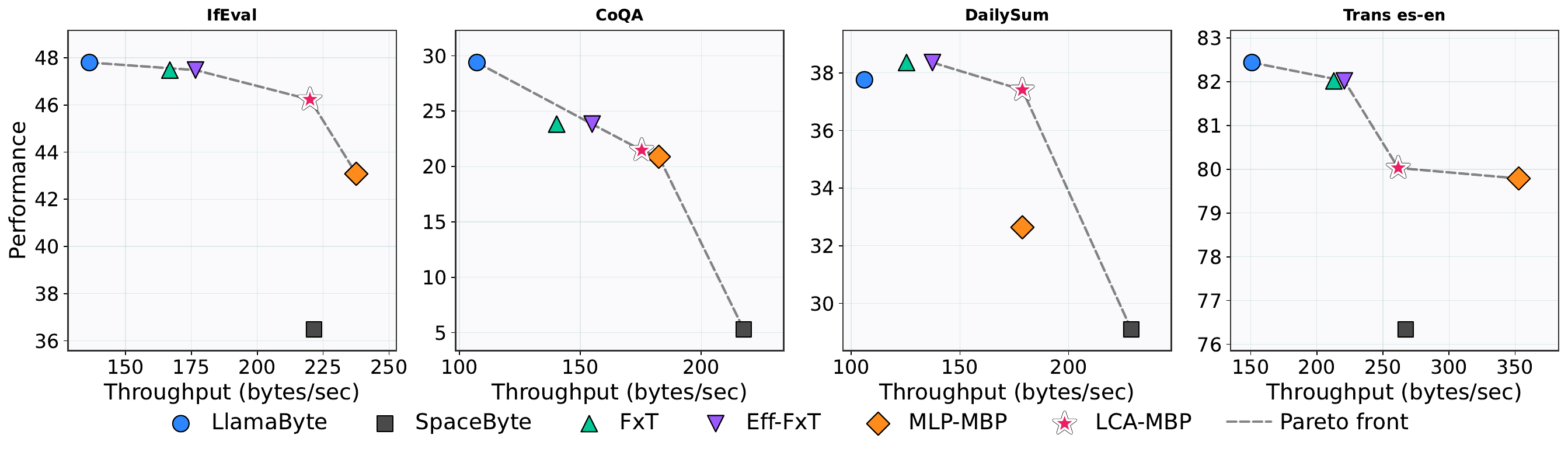}
    \caption{\textbf{Performance vs.\ throughput across four downstream tasks.} Metrics are instruction-loose accuracy on IfEval, F1 on CoQA, ROUGE on CNN/DailyMail, and COMET on Spanish--English translation. \methodname (\textcolor[HTML]{E91E63}{$\star$}) lies on the frontier for three of four tasks on the Pareto front, which represents the method with the best tradeoff.}
    \label{fig:pareto_front_res}
\end{figure*}

\subsection{Model Architecture}
For all models, we use a hidden dimension of $d_{\text{model}} = 1024$, an LM inner dimension of $d_{\text{inner}} = 4096$, and a context window of 4096. We also use RMSNorm with $\epsilon = 10^{-6}$, a rotary embedding base of 100000, $8$ attention heads, $8$ KV heads, dropout of $0.1$, and attention scaling of $1.0$. Unless otherwise stated, we keep these architectural and optimization settings fixed across runs. The main architectural differences across models, therefore, lie in how layers are allocated across the encoder, LM module, next-token (NT) decoder, and multi-byte (MB) decoder. For the plain LlamaByte baseline, all layers are placed in a single stack, while the hierarchical baselines distribute layers across the encoder, LM module, and decoder components. Among these hierarchical variants, the principal difference is the choice of MB prediction head: either no MB head, multiple MLP heads, or a two-layer \methodname head.

\subsection{Model Training}
We follow Llama 3  \citep{grattafiori2024llama} for the general implementation of all parts of our model, and we detail our configurations in \autoref{apx:archi_and_hyperparams}.
We pretrain a 373M-parameter hierarchical transformer model on 50B bytes from FineWeb-edu English dataset \citep{lozhkov2024fineweb-edu}. We train with a total batch size of 1048576 bytes per gradient step and a warmup of 2000 steps. We use a learning rate of 3e-4  and the Adam optimizer with a cosine learning rate scheduler. We set Adam $\beta_1$ and $\beta_2$ to 0.9 and 0.999, respectively, while we set the $\epsilon$  to 1e-8. We employ gradient clipping at 0.25. All models were trained on 4 NVIDIA B200 GPUs.

We pretrain multiple models with different configurations. We use a compression rate of 3$\times$. Following \citep{owodunni2025flexitokens}, we achieve this by setting $\beta$ and $\alpha$ in the BP loss $\mathcal{L}_{BP}$ to 0.261 and 0.333 respectively. 
We use 3 heads for MLP-MBP and train all models for 1 epoch and 
we set $\lambda_0=1$, $\lambda_1=1$ and $\lambda_2=10$.

\subsection{Downstream Fine-tuning and Evaluation}
We perform supervised finetuning on Tulu-3 SFT mix \citep{lambert2024tulu3} for 5 epochs. We use the same hyperparameters as pretraining except for the learning rate, which we set to 2e-4, batch size to 64, $\lambda_1=2$. To evaluate on translation, we separately finetune our pretrained models on the Spanish and French datasets in Opus-100 \citep{zhang-etal-2020-improving} for 2 epochs with the SFT hyperparameters.


We evaluate our models on 4 major generative tasks that effectively measure our speed-to-performance comparison. These include Summarization \citep[CNN/Daily Mail;][]{DBLP:conf/nips/HermannKGEKSB15}, Question Answering \citep[CoQA;][]{reddy-etal-2019-coqa}, Machine translation \citep[Opus-100;][]{zhang-etal-2020-improving}), and Instruction Following \citep[IFEval;][]{zhou2023instructionfollowingevaluationlargelanguage}). At inference time, we use a temperature of 0.7 and  \texttt{top\_p} of 0.9 for IFEval, while we use greedy decoding for other tasks since they are factual tasks where the output should closely match a reference. Unless otherwise stated, we set the acceptance threshold $\tau=0.9$ and candidate bytes $n=3$ for our main experiments. In all our evaluations, we report wall-clock on a single B200 with a batch size of 1.

\section{Results and Analyses}
\label{sec:results}
\autoref{fig:pareto_front_res} compares all six methods across the four downstream tasks. We highlight two main findings.

\methodname-MBP lies on the Pareto front in three of four tasks. Across IFEval, CoQA, and CNN/DailyMail summarization, \methodname matches or exceeds the throughput of other methods while retaining task performance close to the strongest non-MBP hierarchical models (FxT and Eff-FxT). On Spanish--English translation, FxT and Eff-FxT achieve slightly higher COMET scores, but at substantially lower throughput; \methodname trades a small amount of performance for a meaningful throughput gain. Notably, this result holds even though our pretraining corpus (FineWeb-edu) contains no dedicated translation data, suggesting that the multi-byte prediction structure can transfer to non-English generation after lightweight finetuning. Across all four tasks, \methodname achieves this trade-off with a single decoder head.

\autoref{tab:acceptance_rate} reports the overall acceptance rates of our LCA and MLP-based multi-byte prediction models. 
The Pareto comparison in \autoref{fig:pareto_front_res} reveals that despite its high acceptance rate, MLP-MBP consistently underperforms \methodname on downstream metrics at comparable throughput. This indicates that MLP-MBP is highly confident in its speculative predictions, but those predictions are frequently incorrect. 

\insertacceptancerate

We attribute this to the conditional independence assumption made by the MLP heads: each future token is predicted from a shared hidden state without conditioning on the other tokens being predicted in parallel. \methodname, by contrast, uses transformer layers in the MBP head to condition each predicted byte on the previous segment's context and on its position within the current segment, producing more coherent multi-byte predictions than independent MLP heads operating on the same shared hidden state. The result is a lower acceptance rate but higher end-to-end quality---a more useful operating point in practice.

\section{Discussion and Ablations}
\label{sec:analysis}
\subsection{Ablation on the Acceptance Threshold $\tau$}
\begin{figure}[t]
    \centering
    \includegraphics[width=1\linewidth]{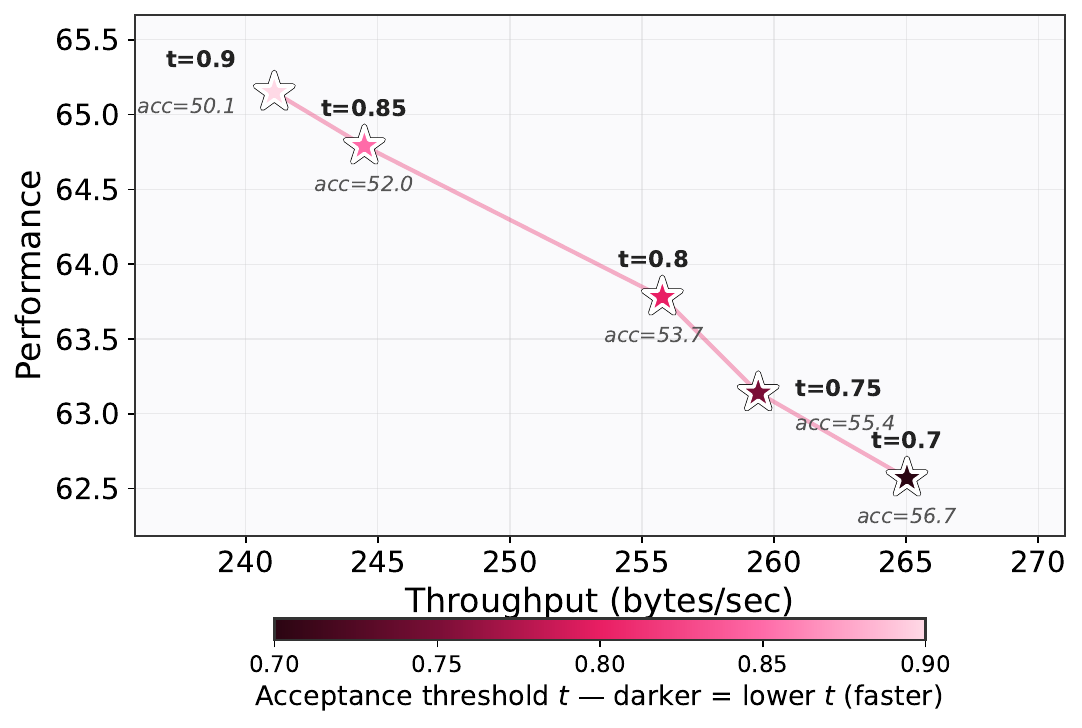}
    \caption{\textbf{Ablation on the acceptance threshold, averaged across DailySum, es-en, and fr-en.} Each star marks \methodname evaluated at a different value of $\tau$. Lowering $\tau$ admits more speculative tokens, increasing throughput and acceptance rate.}
    \label{fig:threshold_ablation}
    \vspace{-0.4cm}
\end{figure}

In \autoref{fig:threshold_ablation}, we report average performance, throughput, and acceptance rate obtained by varying $\tau$ from 0.9 to 0.7, averaged across DailySum, es-en, and fr-en. As $\tau$ decreases, more speculative bytes are admitted: the acceptance rate rises from 50.1\% to 56.7\% and throughput improves by roughly 10\%, while average performance declines by about 3 points. The drop is steep around $\tau=0.75$, where the marginal throughput gain no longer justifies the loss in quality. Based on these results, we therefore can deduce that $\tau=0.75$ offers a balanced trade-off between speed and downstream quality for further ablations.
\subsection{Impact of Increasing Candidate Tokens Beyond Compression Rate}
\begin{figure}[h]
    \centering
    \includegraphics[width=1\linewidth]{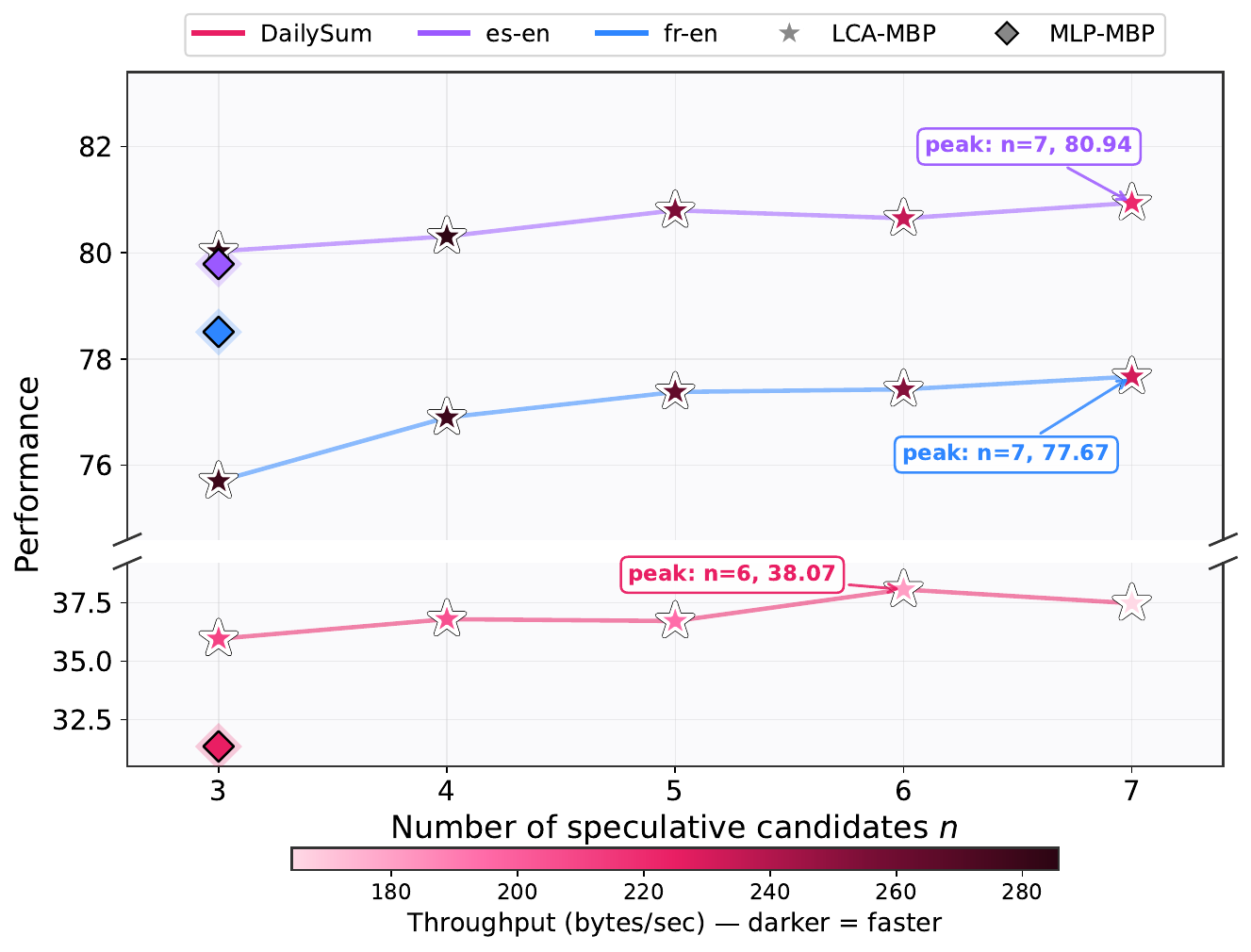}
    \caption{\textbf{Performance vs. Candidates ($n$  at $\tau=0.75$.)} Marker color encodes throughput (darker = faster; full values in Table~\ref{tab:n_ablation_throughput}). COMET scores (es-en, fr-en) scaled $\times$100.}
    \label{fig:n_ablation}
    \vspace{-0.4cm}
\end{figure}

\begin{figure}[h]
    \centering
    \includegraphics[width=1\linewidth]{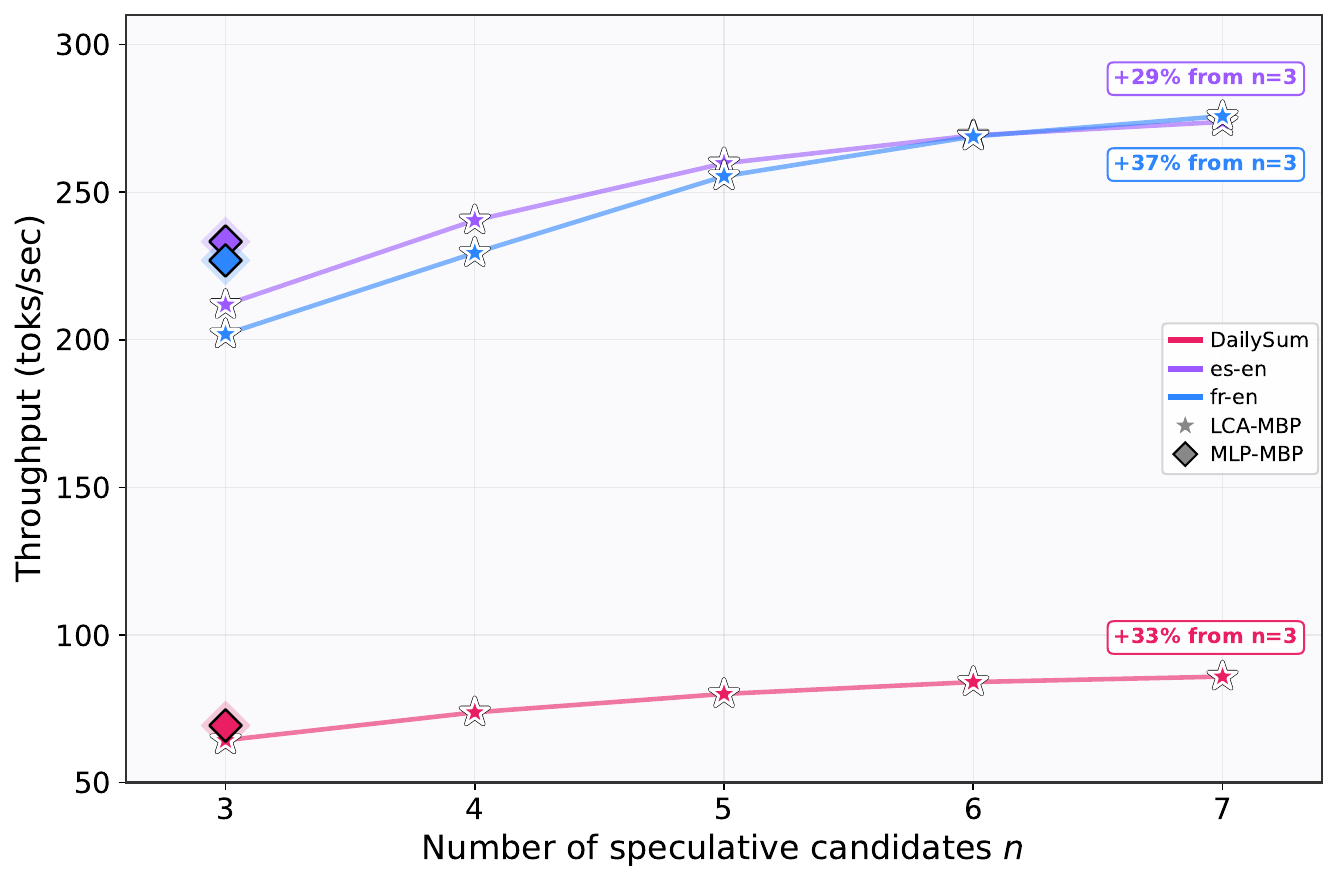}
    \caption{\textbf{Throughput vs. candidates $n$ under speculative decoding acceptance.} Solid lines show \methodname throughput growing monotonically with $n$ on all three tasks while downstream performance is maintained (\autoref{tab:spec_decode_performance} in \autoref{apx:extended_results}).}
\label{fig:n_ablation_verifier}
\end{figure}

Our \methodname-MBP model allows the number of speculative candidates $n$ to be varied at inference time without retraining, by duplicating the last latent token as described in Line 10 of Algorithm \autoref{alg:inference}.  In this subsection, we study how this concept works by evaluating our model on multiple benchmarks using two candidate acceptance methods: \textbf{(i)} \textbf{Probability threshold} acceptance as discussed in Algorithm \autoref{alg:inference}, and \textbf{(ii)} \textbf{speculative decoding} acceptance, where the candidate bytes from the multi-byte prediction head are verified with an extra forward pass through the next-byte prediction head. We employ the speculative decoding method as proposed in \citet{leviathan2023fast}.
\paragraph{Increment with Probability Threshold Acceptance:}
\autoref{fig:n_ablation} shows performance and throughput at $\tau=0.75$ as $n$ varies from 3 to 7. We find that the optimal $n$ can be task-specific: \texttt{es-en} and \texttt{fr-en} peak at $n=7$ with about +3 points gain over $n=3$, while DailySum peaks at $n=6$ with +2.11 points. Beyond the peak, performance degrades as the model is asked to predict further past its trained horizon, and throughput drops monotonically due to falling acceptance rates ( see \autoref{tab:n_ablation_throughput}). 
These results indicate that $n$ can be tuned per task when using the threshold acceptance strategy.
\paragraph{Increment with Speculative Decoding Acceptance:}
Unlike in probability threshold acceptance where increasing $n$ changes both performance and inference speed, speculative decoding acceptance guarantees that all the model's output matches those of decoding with the next-byte prediction head. In \autoref{fig:n_ablation_verifier}, we find that downstream performance is constant across $n$, while throughput grows monotonically, gaining +29-37\% between $n=3$ and $n=7$-8. \methodname also overtakes the MLP-MBP baseline on all three tasks. Unlike MLP-MBP, whose number of prediction heads is fixed at training time, \methodname allows $n$ to be increased freely at inference. Under speculative decoding verification, $n$ therefore acts as a free throughput knob: quality is preserved by construction while decoding speed scales with $n$. 

\begin{figure}[h]
    \centering
    \includegraphics[width=1\linewidth]{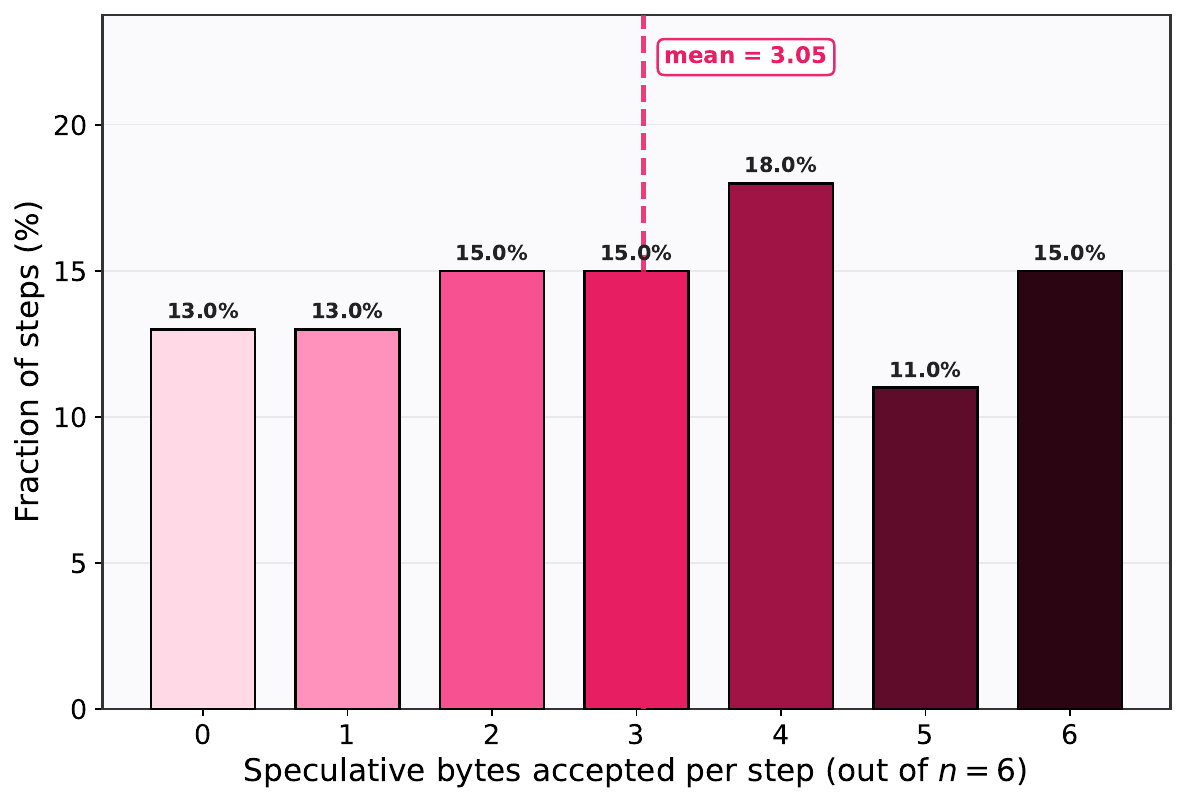}
    \caption{\textbf{Distribution of accepted speculative bytes over 100 decoding steps.} \methodname accepts a mean of 3.05 of the 6 speculative candidates per step, and 15\% of steps accept the full window, indicating that \methodname's multi-byte predictions are frequently coherent across an entire segment rather than accepted one byte at a time.}
\label{fig:draft_acceptance_histogram}
\vspace{-0.4cm}

\end{figure}

\paragraph{Analysis of Accepted Bytes with Increasing Candidates ($n$):} Given the ability to increase the number of candidate bytes in the MBP head beyond the trained compression rate, we investigate what fraction of candidate bytes are accepted from the MBP head alone over 100 decoding steps. \autoref{fig:draft_acceptance_histogram} shows that the number of accepted bytes varies widely from step to step: the model accepts anywhere from 0 to all 6 candidates, with a mean of 3.05 accepted per step and 15\% of steps accepting the entire window. We also observe that no candidate byte is accepted at the first decoding step across multiple runs, as the model has not yet accumulated enough context to make confident multi-byte predictions.

\subsection{Improving Baseline Throughput with \methodname Models}
\label{sec:external_verification}
\begin{figure*}[h]
    \centering
    \includegraphics[width=1\linewidth]{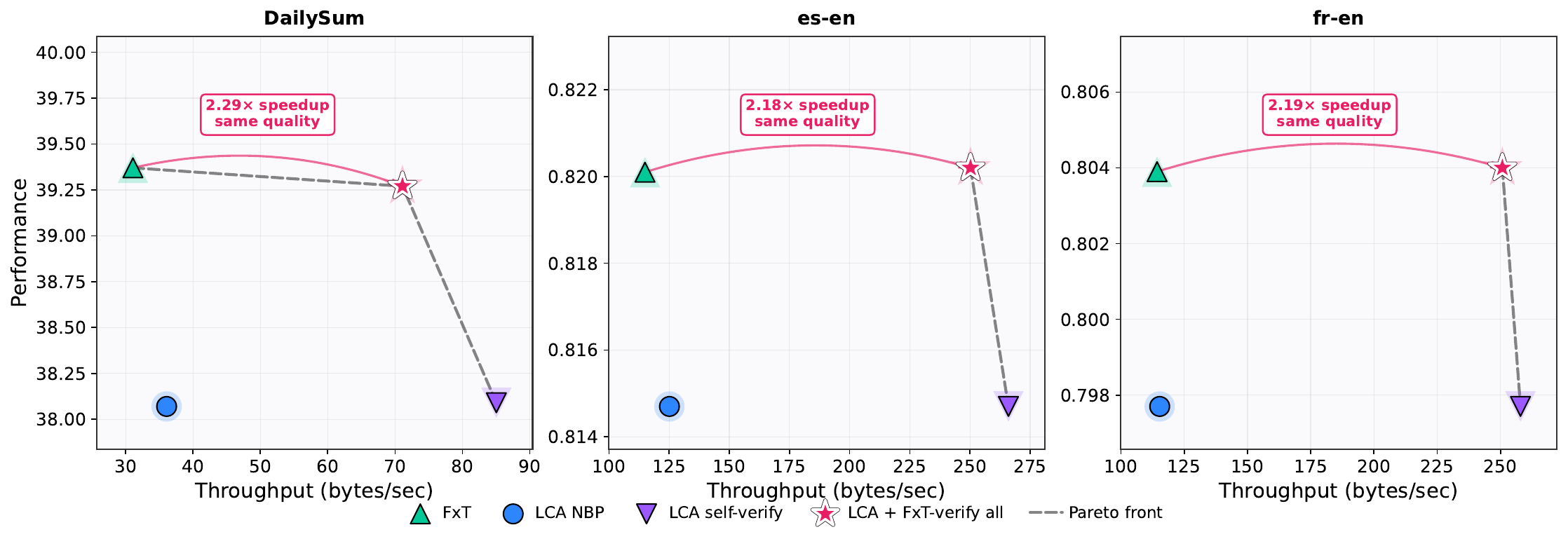}
    \caption{\textbf{Throughput vs.\ performance across four inference strategies.} We compare \methodname's next-byte head only, \methodname's self-verified multi-byte prediction, \methodname with FxT verifying the MBP head, and the FxT baseline. External verification with FxT lets \methodname match or exceed FxT's performance while running 2.1-2.3$\times$ faster across all three tasks with $n=7$. See acceptance rate in \autoref{tab:verify_all_acceptance_rate} of \autoref{apx:extended_results}.}
\label{fig:external_verifier}
\vspace{-0.5cm}
\end{figure*}

Here, we extend our analysis with speculative decoding acceptance to include generating candidate bytes that are then verified by an external baseline model. This ablation investigates how \methodname can be used to accelerate a baseline model by increasing its inference throughput while keeping the performance the same. In \autoref{fig:external_verifier}, where we jointly use our \methodname model as a drafter and an external FxT baseline as the verifier, we observe that
verifying \methodname's multi-byte prediction head with the FxT model matches FxT's downstream performance on all three tasks while achieving a 2.1-2.3$\times$ speedup, even though both models are the same size.

This result highlights \methodname's role as a drop-in accelerator. Notably, the speedup does not come from a smaller draft model, as is typical in speculative decoding \citep{leviathan2023fast, chen2023accelerating}, but from predicting an entire latent token's worth of bytes in parallel. This suggests that any hierarchical byte-level model can be paired with an \methodname counterpart to gain substantial throughput improvements.

\section{Related Work}

\subsection{Hierarchical Models and the Cost of Byte-Level Generation}

Subword tokenization remains the dominant approach for language modeling, but fixed tokenizers introduce well-known limitations for rare words, domain shift, and evolving vocabularies \citep{sennrich-etal-2016-neural}. Byte-level models avoid these issues by operating directly on characters or UTF-8 bytes \citep{xue-etal-2022-byt5, limisiewicz-etal-2024-myte, wang2024mambabyte, AlRfou2018CharacterLevelLM}, but this comes at the cost of substantially longer sequences. As a result, byte-level models face a higher attention cost during training and slower autoregressive generation at inference time.

Recent hierarchical models address this problem by compressing byte or character sequences into shorter latent tokens before applying deeper sequence modeling. Some works explore sequence compression via pooling over fixed-size segments \citep{nawrot-etal-2022-hierarchical, clark-etal-2022-canine, godey-etal-2022-manta, tay2022charformer, yu2023megabyte}, while others employ boundary predictors that dynamically determine where to pool \citep{nawrot2023efficient, hwang2025dynamic}. These boundary predictors are either trained in isolation, as in BLT \citep{pagnoni2025byte}, or jointly optimized with the main language modeling objective \citep{hwang2025dynamic, owodunni2025flexitokens, ahia-etal-2023-languages}. A related line of work studies how controlling the compression rate of a sequence affects performance across multiple languages and domains \citep{ahia2024magnet,owodunni2025flexitokens, limisiewicz2026compute}.

However, most hierarchical models use latent tokenization primarily to reduce sequence length or improve representation learning. At inference, they still follow standard autoregressive decoding, producing one byte at a time. Thus, while hierarchical models reduce the cost of processing byte sequences, they do not address the decoding bottleneck. Our work builds on hierarchical byte-level modeling, but uses learned latent tokens not only as compressed representations but also as the unit for parallel multi-byte prediction. A concurrent work, FastBLT \citep{kallini2026fast}, uses diffusion language modeling to accelerate byte-level inference; unlike our approach, its boundary predictor is trained independently of the language model, as in BLT.
\subsection{Improving Inference Speed with Multi-Token Prediction}
Autoregressive language models generate one token per forward pass, making decoding latency a major bottleneck. Multi-token prediction (MTP) addresses this limitation by training models to predict multiple future tokens from the same context. Prior work has shown that predicting multiple future tokens can improve sample efficiency and accelerate generation \citep{gloeckle2024better, li2024eagle, gerontopoulos2026multi, ankner2024hydra}. In practice, however, most MTP methods rely on additional prediction heads, auxiliary draft modules, or fixed future offsets, while others have not be explored for hierarchical models and multilingual settings \citet{grivas2025fast}.


Medusa \citep{cai2024medusa}, for example, attaches multiple decoding heads to a base model, and similar embedded MTP approaches use separate lightweight heads for each future position \citep{cai2025fastmtp}. These designs share two limitations especially relevant to hierarchical byte-level models: the number of prediction heads grows with the number of future tokens, adding parameter cost, and fixed-offset prediction treats the next $n$ tokens as a static horizon, even though byte sequences have variable local structure and may be better predicted as learned segments.

\methodname addresses both limitations by tying multi-byte prediction to the segments produced by the hierarchical boundary predictor. A single decoder head equipped with the \methodname mask predicts all bytes within a segment in parallel, conditioned only on previous segments. This removes the per-token parameter overhead of multi-head MTP and replaces fixed-offset prediction with variable-length, segment-aligned prediction.

\section{Conclusion}
We introduced \methodname, a dynamic multi-byte prediction method for hierarchical byte-level LMs. Unlike prior MBP approaches that scale parameters with the number of tokens predicted in parallel, \methodname uses a single decoder head and a boundary-aware attention mask to predict a variable number of bytes per step. Our method aligns MBP with the hierarchical structure already present in modern byte-level models. Across four downstream tasks, \methodname lies on the Pareto front of performance and throughput on three of four. We show that latent tokens learned by hierarchical models can be reused as units of parallel generation, removing the need for separate per-token MBP heads.

\section*{Acknowledgments}

This material is partially based upon work supported by OSU Kirwan Institute. We are also grateful to Amazon Research Awards (AWS), Ohio Supercomputing Center (OSC), UFIT Research Computing, and NAIRR Pilot Program (NAIRR250264) for providing the compute resources for this work.

\section*{Limitations}%
First, we conduct experiments at a single model scale (373M parameters) and training corpus, which lets us compare \methodname against parameter-matched baselines under identical conditions; verifying that the gains persist at larger scales remains future work and would require more compute resources that are not available to us. Second, our evaluation covers four English-centric or English-paired tasks and does not include low-resource or morphologically rich languages, though \methodname inherits its tokenization behavior directly from the boundary predictor and should transfer without architectural change. Finally, the speculative horizon $n$ is task-dependent under threshold acceptance, which adds a small tuning cost; however, this cost disappears under speculative decoding verification (\autoref{fig:external_verifier}), where $n$ can be increased without changing model output.

\bibliography{custom}

\appendix
\section*{Appendix}

\section{General Architecture, Training Configurations, and Hyperparameters}
\label{apx:archi_and_hyperparams}

\paragraph{Reading Figure~\ref{fig:mbp_lca_architecture}.}
We use a consistent visual encoding throughout the figure. The uniform red boxes denote the processed bytes sequence. In the boundary predictor, multi-colored bands indicate tokenization into segments, with bytes sharing a color belonging to the same segment. Deeper shades (green, orange, blue, yellow) mark latent tokens, which are pooled aggregates of the bytes in each segment, and rotated color blocks indicate positions where rotary positional embeddings (RoPE) have been applied. Broken outlines on the latent tokens after the LM module signal that those tokens have been contextualized by the LM transformer stack. Finally, module backgrounds encode the attention mask used by each module: yellow for standard causal attention and orange for the \methodname mask.

\subsection{Hyperparameters}
Table~\ref{tab:hyperparams} summarizes the architectural configurations used for the main byte-level model variants. All models use the same optimizer family and training hyperparameters unless explicitly noted elsewhere. For the MLP-MBP configuration trained at compression rate $5$, the MB decoder depth is increased from $3$ to $5$ layers.
\inserthyperparams

\section{Extended Results}
\label{apx:extended_results}

\begin{figure}[h]
    \centering
    \includegraphics[width=1\linewidth]{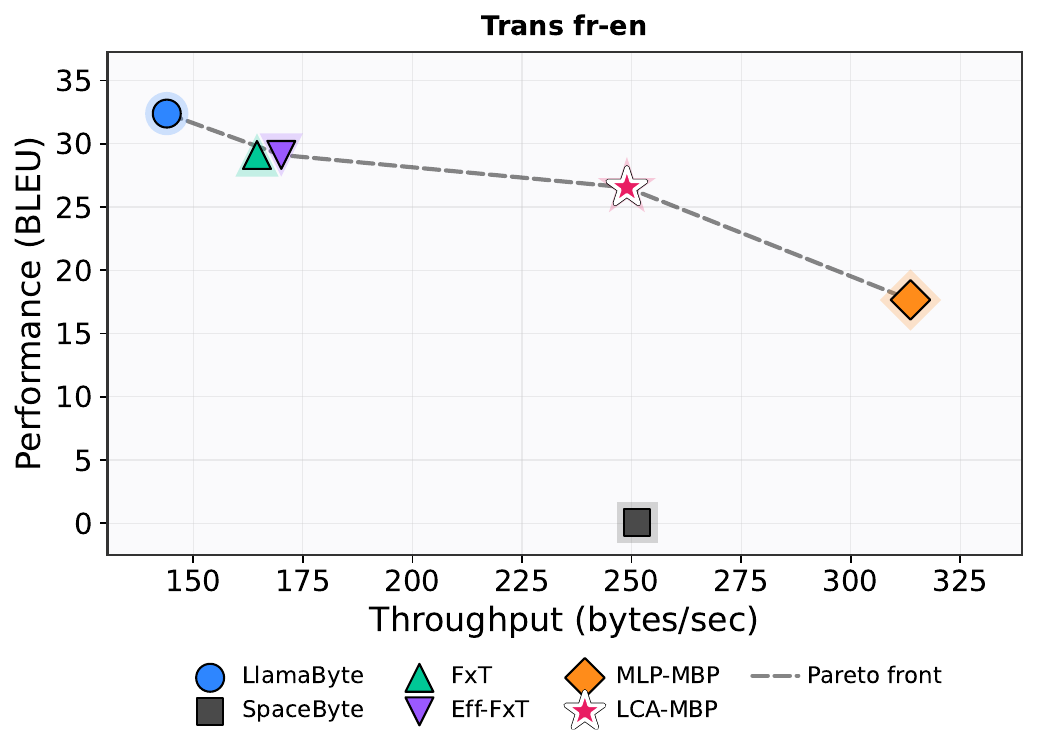}
    \caption{\textbf{Performance vs.\ throughput on French--English translation.} \methodname (\textcolor[HTML]{E91E63}{$\star$}) lies on the Pareto front.}
    \label{fig:pareto_front_fren}
\end{figure}

\paragraph{French Translation Results.}
\autoref{fig:pareto_front_fren} reports performance versus throughput on French$\rightarrow$English translation. \methodname lies on the Pareto front, achieving substantially higher BLEU than MLP-MBP at similar throughput. The trend mirrors the Spanish$\rightarrow$English result in \autoref{fig:pareto_front_res}.

\paragraph{Throughput for Candidate Bytes $n$ Ablation;}
\autoref{tab:n_ablation_throughput} reports throughput (tokens/second) for the $n$-ablation in \autoref{fig:n_ablation}. Throughput decreases monotonically with $n$ on all three tasks, as larger speculative windows incur more rejected candidates per step. The drop is steepest on DailySum (-22\% from $n=3$ to $n=7$), reflecting its lower baseline throughput and tighter sensitivity to rejection overhead.
\insertnablationthroughput

\paragraph{Performance and Acceptance Rate with Speculative Decoding:} Here we report acceptance rate and performance at $n=3$ for our speculative decoding results in \autoref{tab:spec_decode_performance}. We dropped KV caching for all our results that use this method. We note that \methodname-MBP outperforms MLP-MBP on multiple tasks despite achieving a lower acceptance rate.

\insertspecdecodeperf

\paragraph{Throughput and Acceptance rate for LCA with External Verification:}
\autoref{tab:verify_all_acceptance_rate} reports the token acceptance rates for the external verification experiment in \Sref{sec:external_verification}. FxT and LCA NBP accept every token by construction, since they perform standard next-byte decoding without speculation. Under speculative decoding, LCA Self-verify and LCA FxT-verify accept 78-84\% of speculative bytes across the three tasks, meaning that most multi-byte predictions are confirmed on the first pass and only a minority require a fallback to the verifier.
\insertverifyallacceptancerate

\end{document}